\def\year{2020}\relax
\documentclass[letterpaper]{article} 
\usepackage{aaai20}  
\usepackage{times}  
\usepackage{helvet}  
\usepackage{courier} 
\usepackage[hyphens]{url} 
\usepackage{graphicx} 
\usepackage{graphicx} 
\usepackage{epsfig}
\usepackage{amsmath}
\usepackage{amssymb}
\usepackage{amsthm}
\usepackage{algorithm,algorithmic}
\usepackage{subfigure}
\usepackage{amsfonts}
\usepackage{multirow,array,bm}
\usepackage{color}
\usepackage{enumerate}
\usepackage{microtype}      
\usepackage{caption}
\usepackage{epstopdf}
\usepackage{booktabs}  
\usepackage{threeparttable}  

\usepackage{paralist} 

\makeatletter

\newcommand{\Rmnum}[1]{\expandafter\@slowromancap\romannumeral #1@}
\makeatother
\begin{document}
%
\title{GTNet: Generative Transfer Network for Zero-Shot Object Detection}
\author{
    Shizhen Zhao\textsuperscript{\rm 1},
    Changxin Gao\textsuperscript{\rm 1}\thanks{Corresponding author.},
    Yuanjie Shao\textsuperscript{\rm 1},
    Lerenhan Li\textsuperscript{\rm 1},
    Changqian Yu\textsuperscript{\rm 1},
    Zhong Ji\textsuperscript{\rm 2},
    Nong Sang\textsuperscript{\rm 1}\\
    \textsuperscript{\rm 1}Key Laboratory of Ministry of Education for Image Processing and Intelligent Control,\\School of Artificial Intelligence and Automation, Huazhong University of Science and Technology\\
    \textsuperscript{\rm 2}School of Electrical and Information Engineering, Tianjin University\\
{\tt\small Email: \{zhaosz, cgao\}@hust.edu.cn}
}
\maketitle

\begin{abstract}

We propose a Generative Transfer Network (GTNet) for zero-shot object detection (ZSD).
GTNet consists of an Object Detection Module and a Knowledge Transfer Module.
The Object Detection Module can learn large-scale seen domain knowledge.
The Knowledge Transfer Module leverages a feature synthesizer to generate unseen class features, which are applied to train a new classification layer for the Object Detection Module. 
In order to synthesize features for each unseen class with both the intra-class variance and the IoU variance, we design an IoU-Aware Generative Adversarial Network (IoUGAN) as the feature synthesizer, which can be easily integrated into GTNet. 
Specifically, IoUGAN consists of three unit models: Class Feature Generating Unit (CFU), Foreground Feature Generating Unit (FFU), and Background Feature Generating Unit (BFU).  
CFU generates unseen features with the intra-class variance conditioned on the class semantic embeddings. 
FFU and BFU add the IoU variance to the results of CFU, yielding class-specific foreground and background features, respectively.
We evaluate our method on three public datasets and the results demonstrate that our method performs favorably against the state-of-the-art ZSD approaches. 

\end{abstract}

\section{1. Introduction}
\label{sec_intro}
In recent years, many deep learning methods have achieved desirable performance in object detection~\cite{Girshick_2014,Girshick_2015,Ren_2015,Redmon_2016,Liu_2016,he2017mask}.
However, the performance of the detectors relies on the large-scale detection datasets with fully-annotated bounding boxes.
It is impractical to collect enough labeled data since the real world is overwhelmed with an enormous amount of categories.
In this case, these methods are challenged by the task of zero-shot object detection (ZSD), which aims to simultaneously classify and localize new classes in the absence of any training instances.

Zero-shot object detection can be addressed in two spaces: the semantic embedding space and the visual feature space.
(1) Existing methods~\cite{Bansal_2018,Rahman_2018,Li_2019zeroshot} generally map the visual features from predicted bounding boxes to the semantic embedding space.
During the inference stage, a class label is predicted by finding the nearest class based on the similarity scores with all class embeddings.
However, mapping high-dimensional visual feature to low-dimensional semantic space tends to cause the hubness problem due to the heterogeneity gap between these two spaces~\cite{Zhang_2018}. 
(2) Directly classifying an object in the visual feature space can address the hubness problem. 
A number of zero-shot classification methods~\cite{Xian_2018,Verma_2018,li_2019leveraging,huang_2018generative}
have proved the effectiveness of the solution in the visual space. 
However, visual features contain not only the intra-class variance but also the IoU variance, which is a critical cue for object detection. 

To address these problems, we propose a Generative Transfer Network (GTNet) for ZSD.
Specifically, we introduce a generative model to synthesize the visual features to solve the hubness problem.
Meanwhile, with the consideration of the IoU variance, we design an IoU-Aware Generative Adversarial Network (IoUGAN) to generate visual features with both the intra-class variance and the IoU variance.

\begin{figure}[t!]
	\centering
	\includegraphics[width=0.96\linewidth]{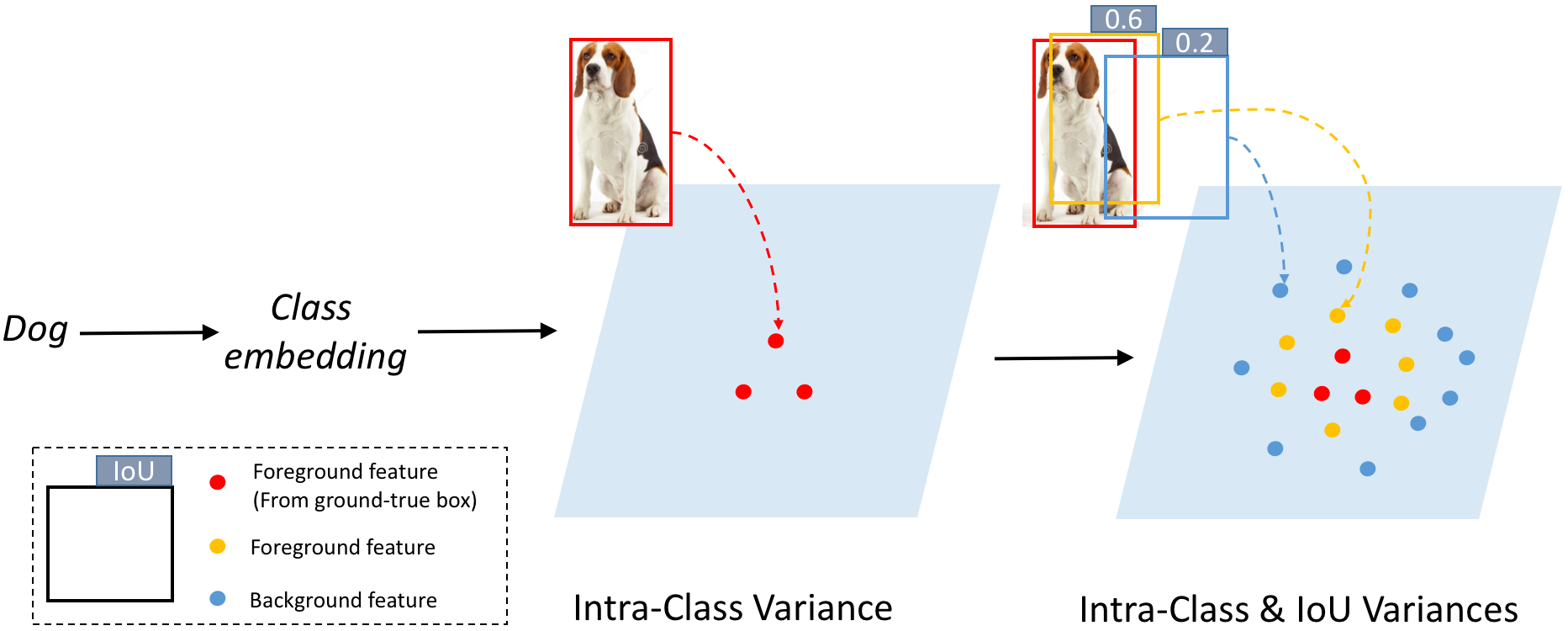}
	\caption{Illustration of our feature generating process. Visual features are first synthesized conditioned on the semantic embeddings. Then we further add the IoU variance to the features. }
	\label{fig:intro}
\end{figure}

The proposed GTNet consists of an Object Detection Module and a Knowledge Transfer Module.
More specifically, the Object Detection Module contains a feature extractor, a bounding box regressor, and a seen category classifier. 
In addition, the feature extractor is used to extract features of Region-of-Interest (RoI) from an image.
The Knowledge Transfer Module consists of a feature synthesizer and an unseen category classifier.
The feature synthesizer is utilized to generate visual features for training the unseen category classifier.
The trained unseen category classifier can be integrated with the feature extractor and the bounding box regressor to achieve ZSD.

As a detection-specific synthesizer, IoUGAN consists of three unit models: Class Feature Generating Unit (CFU), Foreground Feature Generating Unit (FFU), and Background Feature Generating Unit (BFU).  
Each unit contains a generator and a discriminator. 
The feature generating process of IoUGAN is shown as Figure~\ref{fig:intro}.
Specifically, CFU focuses on synthesizing features for each unseen class with intra-class variance conditioned on the class semantic embeddings.
FFU aims to add the IoU variance to the CFU results, generating the foreground features.
Additionally, in order to reduce the confusion between background and unseen categories, BFU synthesizes the class-specific background features conditioned on the CFU results.
We conduct extensive experiments on three public datasets, and those experimental results show that the proposed approach performs favorably against the state-of-the-art ZSD algorithms.

The main contributions of this work can be summarized as follows:
\begin{compactitem}
  \item We propose a novel deep architecture GTNet for the ZSD problem.
  GTNet utilizes a feature synthesizer to generate unseen class features.
  It is only required to train a new classification layer for a pre-trained detector using the synthesized features.
  To the best of our knowledge, we are the first to propose a generative approach for ZSD. 
  
  \item We propose a novel conditional generative model IoUGAN to synthesize unseen class features with both the intra-class variance and the IoU variance. 
  %
  %
  Moreover, IoUGAN can be integrated into GTNet as the feature synthesizer.
  \item We conduct extensive experiments and component studies to demonstrate the superiority of the proposed approach for ZSD. 
\end{compactitem}
\section{2. Related Work}

\noindent \textbf{Fully-Supervised Object Detection:} 
In the past years, object detection is driven by deep convolutional neural networks (CNN). 
The most popular models can be categorized into single-stage networks like SSD~\cite{Liu_2016}, YOLO~\cite{Redmon_2017} and double-stage networks like Faster R-CNN~\cite{Ren_2015}, R-FCN~\cite{dai2016rfcn}. 
Specifically, single-stage networks like SSD~\cite{Liu_2016}, YOLO~\cite{Redmon_2017} take classification and bounding box regression in one step. 
While double-stage networks like Faster R-CNN~\cite{Ren_2015}, R-FCN~\cite{dai2016rfcn} predict offsets of predefined anchor boxes in the first stage by Region Proposal Network (RPN) and then classify and finetune the bounding box prediction in the second stage. 
These object detection models can only detect categories that have appeared in the training dataset. However, they can not be directly applied to predict classes which are unseen during training.

\noindent \textbf{Zero-Shot Classification:} 
Most existing zero-shot classification methods project visual features to the semantic space, such as ~\cite{Frome_2013,Xian_2016,Akata_2015,Kodirov_2017}. Instead of learning a visual-semantic mapping, some previous works ~\cite{Shigeto_2015,dinu2014improving,Zhang_2017} propose to learn a semantic-visual mapping. 
Additionally, there are also some works to learn an intermediate space, which is shared by the visual features and the semantic embeddings~\cite{Changpinyo_2016,Zhang_2016}. 
In contrast, some conditional generative models ~\cite{Xian_2018,Verma_2018,li_2019leveraging,huang_2018generative} transform the zero-shot classification problem to a general fully-supervised problem by leveraging GANs to synthesizing visual features of unseen categories. We also employ a generative model to generate pseudo features for converting the ZSD into a fully supervised object detection problem.
Different the cGANs used in zero-shot classification, IoUGAN can add the IoU variance to the synthesized features. 


\noindent \textbf{Zero-Shot Object Detection:} There are five contemporary studies on ZSD~\cite{Bansal_2018,Rahman_2018,demirel2018zeroshot,Li_2019zeroshot,rahman2018polarity}. 
Specifically, ~\citeauthor{Bansal_2018} proposes a background-aware model to reduce the confusion between background and unseen classes. 
~\citeauthor{Rahman_2018} propose a classification loss composing of a max-margin loss and a meta-class clustering loss.
~\citeauthor{demirel2018zeroshot} uses a convex combination of embeddings to solve the ZSD problem. 
~\citeauthor{Li_2019zeroshot} detects unseen classes by exploring their natural language description.  
~\citeauthor{rahman2018polarity} proposes a polarity loss function for better aligning visual features and semantic embeddings. 
All of these methods focus on mapping the visual features from predicted bounding boxes to the semantic embedding space. 
In contrast, we propose a generative approach to tackle the ZSD problem. 


\section{3. Generative Approach for Zero-Shot Object Detection}

\noindent In this section, we first present GTNet, which consists of two modules: an Object Detection Module and a Knowledge Transfer Module. 
Then we design a novel conditional generative model (\textit{i.e.} IoUGAN) as the feature synthesizer, which is embedded in the Knowledge Transfer Module.  

\begin{figure}[tb]
\begin{center}
\includegraphics[width=0.96\linewidth]{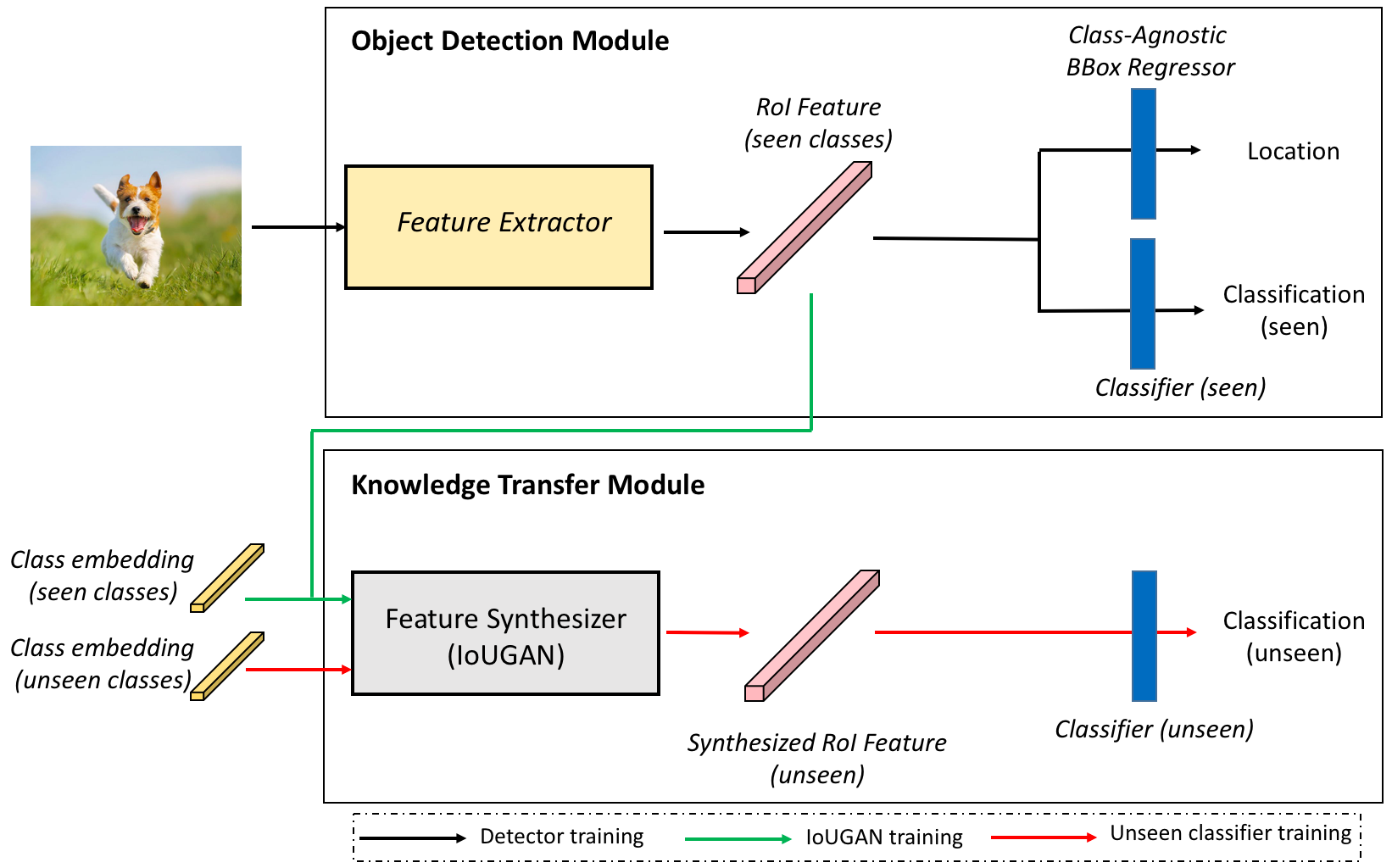}
   \caption{Illustration of the Generative Transfer Network (GTNet). It consists of an Object Detection Module and a Knowledge Transfer Module. 
   The Object Detection Module contains a bounding box regressor, a seen category classifier, and a feature extractor.
   The Knowledge Transfer Module is composed of a feature synthesizer (\textit{i.e.} IoUGAN) and an unseen category classifier.
   The synthesizer can be trained using the features from the feature extractor and the class semantic embeddings.
   A trained synthesizer can sample the unseen class features that can be  applied to train an unseen category classifier. 
   The trained unseen classifier can further be integrated with the feature extractor and the regressor to achieve zero-shot object detection.
   }
 \label{fig:framework}
\end{center}
\end{figure}

\subsection{3.1. Generative Transfer Network}
\label{framework section}
\paragraph{Network Overview:}
As shown in Figure~\ref{fig:framework}, GTNet consists of an Object Detection Module and a Knowledge Transfer Module. 
Specifically, the Object Detection Module is composed of a seen categories classifier, a bounding box regressor, and a feature extractor.
The bounding box regressor is shared among all categories instead of being specific for each category, in order to reuse the regression parameters for detecting unseen classes.
The feature extractor is used to extract RoI features from an image.
Specifically, the feature extractor is in the fashion of Faster R-CNN~\cite{Ren_2015}, which has superior performance among competitive end-to-end detection models.
The feature synthesizer (\textit{i.e.} IoUGAN) is a conditional generative model, which can learn to generate visual features conditioned on the class semantic embeddings. 
IoUGAN will be detailed in Section 3.2.


\paragraph{Building Process of the Zero-Shot Detector:}
The Object Detection Module is pre-trained in the large-sclae seen class dataset. 
Then the feature synthesizer can be trained using the real features from the feature extractor and the corresponding class embeddings.
The trained synthesizer can generate the visual features of unseen classes by inputting the corresponding class semantic embeddings.
To the end, we use the generated features to train an unseen categories classifier. 
In the inference stage, we integrate the trained unseen category classifier with the feature extractor and the regressor to detect unseen classes. 

\begin{figure*}[tb]
\begin{center}
\includegraphics[width=0.9\linewidth]{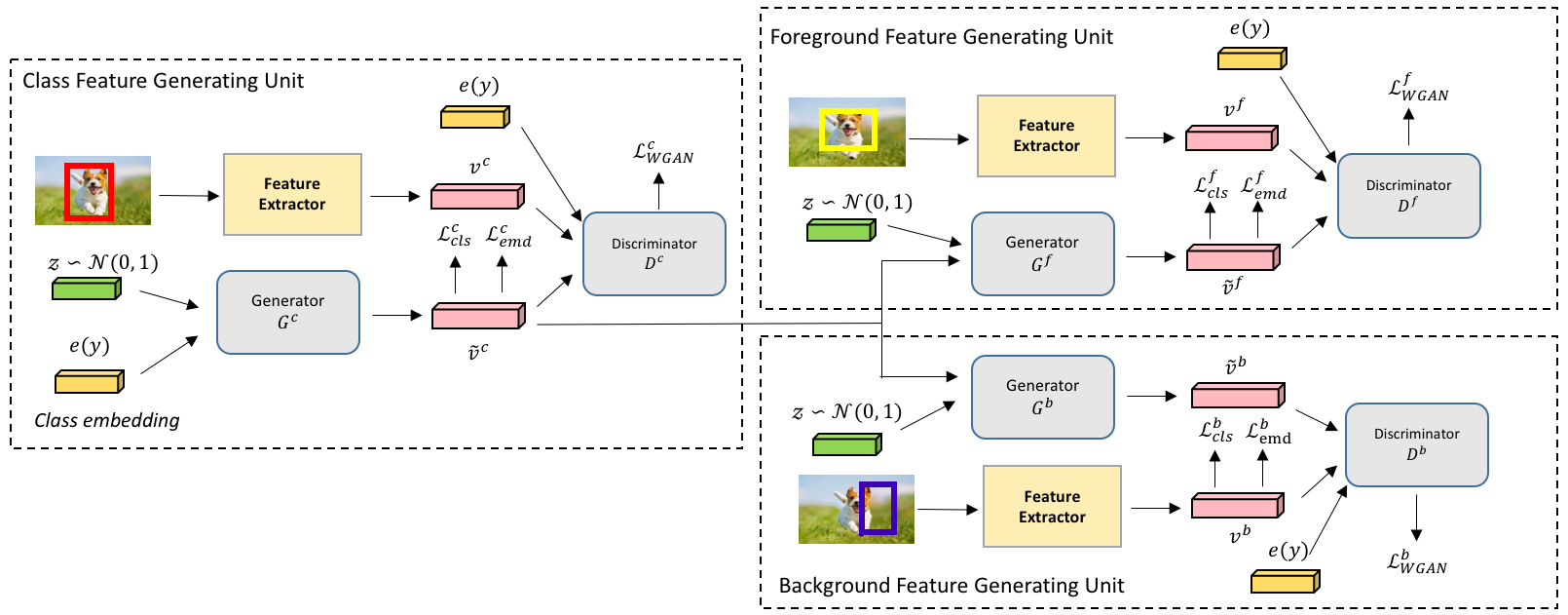}
   \caption{Illustration of the IoU-Aware Generative Adversarial Network (IoUGAN). 
   The Class Feature Generating Unit (CFU) takes the class embeddings and the random noise vectors as input and outputs the features with the intra-class variance. 
   Then the Foreground Feature Generating Unit (FFU) and the Background Feature Generating Unit (BFU) add the IoU variance to the results of CFU and output the class-specific foreground and background features, respectively. 
   }
 \label{fig:IoUGAN}
\end{center}
\end{figure*}

\subsection{3.2. IoU-Aware Generative Adversarial Network}
\label{sec_iougan}
To generate features of different classes with the IoU variance, we propose a simple yet effective IoUGAN, which mainly is composed of three unit models (\textit{i.e.} CFU, FFU, and BFU), as shown in Figure~\ref{fig:IoUGAN}. 
Each unit contains a generator and a discriminator. 
In addition, we use WGAN~\cite{arjovsky2017} as the basic component for the proposed generative model due to its more stable training. 

\begin{compactitem}
\item [-] {\bf CFU}: it focuses on generating features with the intra-class variance conditioned on the given semantic embeddings and the noise vectors. 

\item [-] {\bf FFU}: it adds the IoU variance to the results of CFU and outputs the class-specific foreground features. 

\item [-] {\bf BFU}: it takes the results of CFU as the inputs and outputs the class-specific background features.

\end{compactitem}

\paragraph{Definitions and Notations:}
We use the feature extractor to sample the visual features of seen categories. 
The training set for IoUGAN is denoted as  $S = \{(v_i^c,\{v_{ij}^f, v_{ij}^b\}^{n^s}_{j=1}, y_i, e(y)_i) \}^{N^s}_{i=1}$, 
where $v_i^c \in \mathcal V^c$ denotes the feature extracted from a ground-true box, 
$v_{ij}^{f} \in \mathcal V^{f}$ is the class-specific foreground feature from a positive bounding box whose IoU with corresponding the ground-truth box is larger than a threshold $t_{f}$, 
$v_{ij}^b \in \mathcal V^{b}$ is the class-specific background feature from a negative bounding box whose IoU with the corresponding ground-truth box is smaller than a threshold $t_{b}$, 
$y \in \mathcal Y^s $ denotes the class label in $\mathcal Y^s = \{y_1,\ldots,y_s\}$ with $S$ seen classes and $e(y) \in \mathcal{E}$ denotes the category-specific semantic embedding. 
Note that the positive boxes and the negative boxes, from which $v_{ij}^{f}$ and $v_{ij}^{b}$ are extracted, are predicted by RPN. 
During training, we randomly pick $v_{ij}^{f}$ and $v_{ij}^{b}$ among their $n^s$ samples with the corresponding $v_i^{c}$ to IoUGAN. 
Additionally, we have the semantic embeddings of unseen classes $\mathcal{U} = \{(u,e(u)| u \in \mathcal{Y}^u, e(u) \in \mathcal{E}\}$, where $u$ is a class from a disjoint label set $\mathcal{Y}^u = \{u_1,\ldots,u_u\}$ of $U$ labels and $e(u)$ is the semantic embedding of a corresponding unseen class.

\paragraph{CFU:}
Instead of directly generating features of different class with the IoU variance, we simplify the task to first generate features with the intra-class variance using our CFU. 
We use the feature $v^{c}$ extracted from the ground-truth bounding box as the real feature to guide the generator to capture the overall characteristic of an object. 
Given the train data $\mathcal{S}$ of seen classes, we aim to learn a conditional generator $G^{c}:\mathcal{Z} \times \mathcal{E}\rightarrow \mathcal V^{c}$ in CFU, which takes random Gaussian noise $z\in \mathcal{Z} \subset \mathbb{R}^{d_z}$ and class embedding $e(y) \in \mathcal{E}$ as its inputs, and outputs a visual feature $\tilde{v}^{c} \in \mathcal V^{c}$ of class $y$.
The discriminator $D^{c}: \mathcal V^{c} \times \mathcal{E} \rightarrow \mathbb{R}$ is a multi-layer perceptron which outputs a real value. 
$G^{c}$ tries to minimize it, while $D^{c}$ seeks to maximize the loss.
Once the generator $G^{c}$ learns to synthesize visual features of seen categories, i.e. $v^{c}$, conditioned on the seen class embedding $e(y) \in \mathcal{Y}^s$, it can also generate $\tilde{v}^{c}$ of any unseen class $u$ via its class embedding $e(u)$. 
The conditional WGAN loss in CFU is given by
\begin{equation}
    \mathcal{L}_{WGAN}^{c} = \mathbb{E}[D^{c}(v^{c},e(y))] - \mathbb{E}[D^{c}(G^{c}(z,e(y)),e(y))] -  
\label{gan1}
\end{equation}
\begin{equation*}
    \quad \quad \quad \alpha^c\mathbb{E}[(||\nabla_{\hat{v}^{c}}D(\hat{v}^{c},e(y))||_2 - 1)^2]
\end{equation*}
where $\hat{v}^{c}$ is the convex combination of $v^{c}$ and $\tilde{v}^{c}$: $\hat{v}^{c}=\eta^c  v^{c} + (1-\eta^c) \tilde{v}^{c}$ with $\eta^c \sim U(0,1)$,  $\alpha^c$ is the penalty coefficient and $\mathbb{E}$ is the expectation. 
The Wasserstein distance is approximated by the first two terms, and the third term constrains the gradient of the discriminator to have unit norm along with the convex combination of real (\textit{i.e.}$v^{c}$) and generated  (\textit{i.e.}$\tilde{v}^{c}$) pairs. 

\paragraph{FFU:}
The bounding boxes predicted by RPN always can not overlap entirely with the ground-truth bounding boxes. 
Even a bounding box is a positive one that has large IoU with a ground-truth box, the feature extracted from the box still lacks some information compared to that extracted from a ground-true box. 
The detector should be robust to the information loss of a foreground feature. 
In this case, we use FFU to randomly add the IoU variance to the features outputted by CFU.
Therefore, the foreground feature $v^{f}$ is used as the real feature to train FFU.

With the output feature from CFU: $\tilde{v}^{c}=G^{c}(z,e(y))$ and Gaussian latent variable $z$, the conditional WGAN loss of FFU is given by
\begin{equation}
    \mathcal{L}_{WGAN}^f = \mathbb{E}[D^{f}(v^{f},e(y))] - \mathbb{E}[D^{f}(G^{f}(z,\tilde{v}^{c}),e(y))] -  
\label{gan2}
\end{equation}
\begin{equation*}
    \quad \quad \quad \alpha^f\mathbb{E}[(||\nabla_{\hat{v}^{f}}D^{f}(\hat{v}^{f},e(y))||_2 - 1)^2]
\end{equation*}

\noindent where $\hat{v}^{f}$ is a convex combination of $v^{f}$ and $\tilde{v}^{f}$: $\hat{v}^{f}=\eta^f  v^{f} + (1-\eta^f) \tilde{v}^{f}$ with $\eta^f \sim U(0,1)$, $\alpha^f$ is the penalty coefficient.
Different from the generator in CFU, the class semantic embedding is not used as an input to the generator of FFU with the assumption that the semantic information has already been preserved by $\tilde{v}^{c}$.

\paragraph{BFU:}
Confusion between background and unseen classes limits the performance of a zero-shot detector~\cite{rahman2018polarity,Bansal_2018}. 
To enhance the discrimination of detector between background and unseen classes, we use BFU to generate the class-specific background features. We use the background feature $v^{b}$ as the real feature to train BFU.

With the output feature from CFU: $\tilde{v}^{c}=G^{c}(z,e(y))$ and Gaussian latent variable $z$, the conditional WGAN loss of BFU is given by
\begin{equation}
    \mathcal{L}_{WGAN}^{b} = \mathbb{E}[D_{b}(v^{b},e(y))] - \mathbb{E}[D^{b}(G_{b}(z,\tilde{v}^{c}),e(y))] -  
\label{gan3}
\end{equation}
\begin{equation*}
    \quad \quad \quad \alpha^b\mathbb{E}[(||\nabla_{\hat{v}^{b}}D^b(\hat{v}^{b},e(y))||_2 - 1)^2]
\end{equation*}
\noindent where $\hat{v}^{b}$ is a convex combination of $v^{b}$ and $\tilde{v}^{b}$: $\hat{v}^{b}=\eta^b  v^{b} + (1-\eta^b) \tilde{v}^{b}$ with $\eta^b \sim U(0,1)$, $\alpha^b$ is the penalty coefficient.

\paragraph{Overall Objective Function:}
Additionally, the synthesized features should be well suited for training a discriminative classification layer of a detector. 
Similar to~\cite{Xian_2018}, a discriminative classification layer trained on the input data is used to classify the synthesized features $\tilde{v}$ so that we can minimize the classification loss over the generated features.
Note that for simplicity, we use the $\tilde{v}$ to denote all the generated features (\textit{i.e.} $\tilde{v}^{c}$, $\tilde{v}^{f}$ and $\tilde{v}^{b}$). To this end, e use the negative log likelihood, 
\begin{align}
\mathcal{L}_{cls} = -E_{\tilde{v}\sim p_{\tilde{v}}}[\log P(y| \tilde{v}; \theta)], 
\end{align}
\noindent where $y$ denotes the class label of $\tilde{v}$, $P(y| \tilde{v}; \theta)$ is the predicted probability of $\tilde{v}$ belonging to its true class label $y$. In addition, we further use the $\mathcal{L}_{cls}^{c}$, $\mathcal{L}_{cls}^{f}$,  $\mathcal{L}_{cls}^{f}$ to denote the classification loss functions for the three units, respectively. 
The conditional probability is computed by a single-layer FC network parameterized by $\theta$, which is pretrained on the real features of seen classes. 

Moreover, we expect generated features of a class $y_i$ to be close to the real features of the same class and far from the features of other classes$y_j$ (for $j\neq i$).
We generate matched (same classes) and unmatched (different classes) pairs by pairing the real and generated features in a batch.  
Inspired by ~\cite{Mandal_2019_CVPR}, the distance between the matched and unmatched features can be minimized and maximized, respectively, by the cosine embedding loss, which is given by, 
\begin{equation}
\mathcal{L}_{emb} = \mathbb{E}_{m}[1-cos(v,\tilde{v})] + \mathbb{E}_{um}[\max(0,cos(v,\tilde{v}))]
\end{equation}
where $\mathbb{E}_{m}$  and $\mathbb{E}_{um}$ are the expectations over the matched ($m$) and unmatched ($um$) pair distributions, respectively.
Similarly, we use ${v}$ to denote all the real features (\textit{i.e.} $v^{c}$, $v^{f}$ and $v^{b}$). 
We further use $\mathcal{L}_{emb}^{c}$, $\mathcal{L}_{emb}^{f}$,  $\mathcal{L}_{emb}^{b}$ to denote the embedding loss functions for the three units, respectively. 
While the other losses ($\mathcal{L}_{WGAN}$ and $\mathcal{L}_{cls}$) focus on the similarity between the generated features and the real ones of the same classes, the embedding loss $\mathcal{L}_{emb}$ also emphasizes the dissimilarity between the generated features and the other class ones. 
The final objective for training CFU, FFU and BFU, using $\beta_1$, $\beta_2$, $\beta_3$, $\gamma_1$, $\gamma_2$ and $\gamma_3$ as hyper-parameters for weighting the respective losses, is given by 
\begin{equation}
    \min_{G^{c}} \max_{D^{c}} \mathcal{L}_{WGAN}^{c} + \beta_1 \mathcal{L}_{cls}^{c} + \gamma_1 \mathcal{L}_{emb}^{c} ,
\label{eqn_final_obj1}
\end{equation}

\begin{equation}
    \min_{G^{f}} \max_{D^{f}} \mathcal{L}_{WGAN}^{f} + \beta_2 \mathcal{L}_{cls}^{f} + \gamma_2 \mathcal{L}_{emb}^{f} ,
\label{eqn_final_obj2}
\end{equation}

\begin{equation}
    \min_{G^{b}} \max_{D^{b}} \mathcal{L}_{WGAN}^{b} + \beta_3 \mathcal{L}_{cls}^{b} + \gamma_3 \mathcal{L}_{emb}^{b}.
\label{eqn_final_obj3}
\end{equation}

\begin{table*}[tb]
\fontsize{10}{15} \selectfont  
  \begin{center}
  \caption{Comparisons with the states-of-the-arts on the ILSVRC-2017 dataset. 
  We use mean average precision (mAP) as the evalution metric and present the results in percentages.
  }
  \label{table:imagenet}
  \resizebox{\linewidth}{!} 
  {
  \begin{tabular}{c ccccc ccccc ccccc ccccc ccccc}
  \toprule
  {}   & \rotatebox{90}{p.box} & \rotatebox{90}{syringe} & \rotatebox{90}{harmonica} & \rotatebox{90}{maraca}  & \rotatebox{90}{burrito}  & \rotatebox{90}{pineapple}   & \rotatebox{90}{electric-fan}  & \rotatebox{90}{iPod}  & \rotatebox{90}{dishwater}  & \rotatebox{90}{can-opener} & \rotatebox{90}{plate-rack} & \rotatebox{90}{bench} & \rotatebox{90}{bow-tie} & \rotatebox{90}{s.trunks} & \rotatebox{90}{scorpion}  & \rotatebox{90}{snail}   & \rotatebox{90}{hamster}  & \rotatebox{90}{tiger}  & \rotatebox{90}{ray} & \rotatebox{90}{train} & \rotatebox{90}{unicycle}& \rotatebox{90}{g.ball} & \rotatebox{90}{h.bar} & \rotatebox{90}{\bf mAP } \\
  \midrule
  SAN~\cite{Rahman_2018} & 5.9 & 1.5 & 0.3 & 0.2 & 40.6 & 2.9 & 7.7 & 28.5 & 13.3 & 5.1 & 7.8 & 5.2 & 2.6 & 4.6 & 68.9 & 6.3 & 53.8 & 77.6 & 21.9 & 55.2 & 21.5 & 31.2 & 5.3 & 20.3 \\
  SB~\cite{Bansal_2018} & 6.8 & 1.8 & 0.8 & 0.5 & 43.7 & 3.8 & 8.3 & 30.9 & 15.2 & 6.3 & 8.4 & 6.8 & 3.7 & 6.1 & 71.2 & 7.2 & 58.4 & 79.4 & 23.2 & 58.3 & 23.9 & 34.8 & 6.5 & 22.0 \\
  DSES~\cite{Bansal_2018} & 7.4 & 2.3 & 1.1 & 0.6 & 46.2 & 4.3 & 8.7 & 32.7 & 14.6 & 6.9 & 9.1 & 7.4 & 4.9 & 6.9 & 73.4 & 7.8 & 56.8 & 80.8 & 24.5 & 59.9 & 25.4 & 33.1 & 7.6 & 22.7 \\
  LAB~\cite{Bansal_2018} & 6.5 & 1.6 & 0.7 & 0.5 & 44.1 & 3.6 & 8.2 & 30.1 & 14.9 & 6.4 & 8.8 & 6.4 & 4.1 & 4.8 & 69.9 & 6.9 & 57.1 & 80.2 & 23.6 & 58.2 & 25.1 & 35.6 & 7.2 & 21.9 \\
  ZSDTD~\cite{Li_2019zeroshot}& {\bf 7.8} & 3.1 & 1.9 & 1.1 & 49.4 & 4.0 & 9.4 & 35.2 & 14.2 & 8.1 & {\bf 10.6} & 9.0 & {\bf 5.5} & 8.1 & {\bf 73.5} & 8.6 & 57.9 & {\bf 82.3} & {\bf 26.9} & 61.5 & 24.9 & 38.2 & 8.9 & 24.1 \\
  GTNet& 4.4 & {\bf 30.4} & {\bf 2.3} & {\bf 1.2} & {\bf 51.1} & {\bf 5.2} & {\bf 18.5} & {\bf 40.6} & {\bf 18.0} & {\bf 13.1} & 4.7 & {\bf 13.7} & 4.6 & {\bf 19.2} & 69.7 & {\bf 10.2} & {\bf 74.7} & 72.7 & 1.4 & {\bf 65.7} & {\bf 27.1} & {\bf 40.4} & {\bf 9.1} & {\bf 26.0} \\
  \bottomrule
  \end{tabular}
  }
  \end{center}
\end{table*}

\section{4. Experimental Results}

\noindent In this section, we first describe the datasets and settings for evaluation. 
Then we compare our approach with the state-of-the-arts and conduct the ablation studies. 
Finally, we show the qualitative results of our method. 

\subsection{4.1. Dataset Description}

Our proposed framework is evaluated on three challenging datasets. 

\textit{ILSVRC-2017 detection dataset} ~\cite{Russakovsky_2015} contains 200 object categories. 
The categories were carefully selected with variations of object scale, level of image clutterness, average number of object instance, \textit{etc}.
Following~\cite{Rahman_2018}, we select 23 categories as unseen classes and the rest are as seen classes.

\textit{MSCOCO}~\cite{Lin_2014} was introduced for object detection and semantic segmentation tasks, with each image containing multiple object instances with different factors such as clutter, views and many others.
Following ~\cite{Bansal_2018}, training images are selected from the 2014 training set and testing images are randomly sampled from the validation set.
For seen/unseen category split, we use 48 seen classes for training and 17 unseen classes for testing. 

\textit{VisualGenome} (VG)~\cite{Krishna_2017} was collected with a focus on visual relationship understanding. 
This dataset constitutes of 500 object categories with multiple bounding boxes per image. Following ~\cite{Bansal_2018}, images from part-1 of the dataset are used for training and testing images are randomly sampled from part-2. 
For seen/unseen category split, we use 478 classes for training and 130 classes for testing. 

\noindent {\bf Train/Test Split:}  In a zero-shot learning task, no visual instances of unseen classes are allowed during training. 
In terms of the ILSVRC-2017 detection dataset, following~\cite{Rahman_2018}, images that contain any unseen classes are removed from the training set.
The test set is composed of the rest of images from ILSVRC training dataset and images from validation dataset that have at least one unseen class bounding box. 
For MSCOCO, following~\cite{Bansal_2018}, we also remove all the images from the training dataset which contain any unseen class bounding boxes. 
For VG, however, due to a large number of test classes and dense labeling, most images will be eliminated from the training set. 
Therefore, following~\cite{Bansal_2018}, we do not conduct the same procedure for VG. 

\noindent {\bf Evaluation Metric:} For ILSVRC-2017 detection dataset, we use mean average precision (mAP) to evaluate the performance.
Following~\cite{Bansal_2018}, for MSCOCO and VG, Recall@100 is used as the evaluation metric which defined as the recall when only the top 100 proposals (sorted by prediction score) are selected from an image. 
This is because, for large-scale crowd-annotated datasets such as MSCOCO and VG, it is difficult to annotate all bounding boxes for all instances of a category.
mAP is sensitive to missing annotations and it will count such detections as false positive.   

\noindent {\bf Implementation Details:} The feature extractor is in the fashion of Faster R-CNN~\cite{Ren_2015}.  
In addition, we use Resnet-101 as the backbone of the feature extractor.
The seen category classifier and the class-agnostic regressor are both single Fully-Connected (FC) layer. 
On the other hand, for all three units (\textit{i.e} CFU, FFU, and BFU) in IoUGAN, the generators $G$ are three-layer FC networks with an output layer dimension equal to the size of the feature. 
The hidden layers are of size 4096. 
The discriminators $D$ are a two-layer FC network with the output size equal to 1 and a hidden size equal to 4096. 
The new classification layer is a single-layer FC with an input size equal to the feature size and the output size equal to the number of unseen classes plus one (\textit{i.e.} background class). 
We use $\beta_1 =\beta_2=\beta_3=0.01$ and $\gamma_1=\gamma_2=\gamma_3=0.1$ accross all datasets. Following ~\cite{Bansal_2018}, the parameters $t_f$ and $t_b$ are set to 0.5 and 0.2. 
All the modules are trained using the Adam optimizer with a 10${}^{-4}$ learning rate.
 
\noindent {\bf Compared Methods:} We compare the effect of our proposed approach to three recent state-of-the-art methods  ~\cite{Rahman_2018,Bansal_2018,Li_2019zeroshot}. 
In order to fairly compare with the method~\cite{Li_2019zeroshot}, we use the textual description embedding which is trained by fastText~\cite{Joulin_2017} as the class semantic embedding.
The study ~\cite{Li_2019zeroshot}  extends other the other two ZSD methods~\cite{Rahman_2018,Bansal_2018} using the textual description embedding and report the results.
We just copy and paste the results from the paper~\cite{Li_2019zeroshot} in Table~\ref{table:imagenet} and Table~\ref{table:coco_vg}.

\renewcommand{\arraystretch}{0.6} 
\begin{table*}[tb]
  \footnotesize
  \begin{center}
  \caption{Comparisons with the states-of-the-arts on the MSCOCO and Visual Genome (VG) datasets. 
  We use Recall@100 as the evaluation metric and present the results in percentages.
  }
  \label{table:coco_vg}
  %
  \begin{tabular}{l ccc c ccc} 
  \toprule
  {} & \multicolumn{3}{c}{MSCOCO} & &\multicolumn{3}{c}{Visual Genome} \\
  \cline{2-4} \cline{6-8}  
  IoU & 0.4 & 0.5 & 0.6 & &0.4 & 0.5 & 0.6  \\
  \midrule
  SAN~\cite{Rahman_2018} & 35.7 & 26.3 & 14.5 & & 6.8 & 5.9 & 3.1 \\
  SB~\cite{Bansal_2018} & 37.8 & 28.6 & 15.4 &  & 7.2 & 5.6 & 3.4 \\
  DSES~\cite{Bansal_2018} & 42.6 & 31.2 & 16.3 & & 8.4 & 6.3 & 3.3 \\
  LAB~\cite{Bansal_2018} & 35.2 & 22.4 & 12.1 & &8.6 & 6.1 & 3.3 \\
  ZSDTD~\cite{Li_2019zeroshot} & 45.5 & 34.3 & 18.1 & &9.7 & 7.2 & 4.2 \\
  GTNet & {\bf 47.3} & \bf 44.6 & \bf 35.5 & & \bf 14.3 & \bf 11.3 & \bf 8.9
  \\
  \bottomrule
  \end{tabular}
  \end{center}
\end{table*}

\renewcommand{\arraystretch}{0.6} 
\begin{table*}[tb]
  \footnotesize
  \begin{center}
  \caption{Experimental results of different components on the three challenging benchmark datasets.
  }
  \label{compon}

  {
  \begin{tabular}{l ccc c ccc c c} 
  \toprule
  {} & \multicolumn{3}{c}{MSCOCO} & &\multicolumn{3}{c}{Visual Genome} & & {ILSVRC} \\
  \cline{2-4} \cline{6-8} \cline{10-10}
  IoU      & 0.4 & 0.5 & 0.6 & &0.4 & 0.5 & 0.6 & & 0.5 \\
  \midrule
  Baseline & 27.4 & 18.1 & 14.5 & & 5.2 & 3.0 & 2.3 & & 16.3\\
  CFU   & 29.2 & 27.0 & 24.7 & & 4.5 & 3.6 & 2.7 & & 20.1\\
  CFU + FFU 
           & 43.7 & 39.2 & 33.0 & & 9.3 & 7.5 & 5.8 & & 24.5 \\
  CFU + FFU + BFU
           & {\bf 46.2} & {\bf 43.4} & {\bf 34.9} & & {\bf 13.6} & {\bf 10.6} & {\bf 7.9} & & {\bf 25.5} \\
  \bottomrule
  \end{tabular}
  }
  \end{center}
\end{table*}

\renewcommand{\arraystretch}{0.6} 
\begin{table*}[!tb]
  \footnotesize
  \begin{center}
  \caption{Experimental results of different loss functions on the three challenging benchmark datasets.
  }
  \label{loss_exp}

  {
  \begin{tabular}{l ccc c ccc c c} 
  \toprule
  {} & \multicolumn{3}{c}{MSCOCO} & &\multicolumn{3}{c}{Visual Genome} & & {ILSVRC} \\
  \cline{2-4} \cline{6-8} \cline{10-10}   
  IoU      & 0.4 & 0.5 & 0.6 & &0.4 & 0.5 & 0.6 & & 0.5 \\
  \midrule
  IoUGAN
           & 46.2 & 43.4 & 34.9 & & 13.6 & 10.6 & 7.9 & & 25.5 \\
  IoUGAN (+ $\mathcal{L}_{cls}$)  
           & 47.1 & 44.3 & 35.3 & & 14.0 & 10.9 & 8.4 & & 25.7  \\
  IoUGAN (+ $\mathcal{L}_{emb}$)  
           & {\bf 47.5} & {\bf 44.9} & 35.2 & & 13.7 & 10.8 & 8.3 & & 25.8  \\
  IoUGAN (+ $\mathcal{L}_{cls}$ + $\mathcal{L}_{emb}$)
           & 47.3 & 44.6 & {\bf 35.5} & & {\bf 14.3} & {\bf 11.3} & {\bf 8.9} & & {\bf 26.0}  \\
  \bottomrule
  %
  \end{tabular}
  }
  \end{center}
\end{table*}

\renewcommand{\arraystretch}{0.6} 
\begin{table}[!tb]
  \footnotesize
  \begin{center}
  \caption{Experimental results of Generalized Zero-Shot Object Detection on the MSCOCO Dataset.}
  \label{gzsl}

  {
  \begin{tabular}{l cc} 
  \toprule
  {} & \multicolumn{2}{c}{MSCOCO} \\
  \cline{2-3}  
        & Seen & Unseen  \\
  \midrule
  Baseline &20.3 & 8.9\\
  Polarity Loss  & 38.2  & 26.3    \\
  GTNet & {\bf 42.5} & {\bf 30.4}   \\
  \bottomrule
  %

  \end{tabular}
  }

  \end{center}

\end{table}

\subsection{4.2. Comparisons with the State-of-the-Arts}
We compare the proposed approach with the state-of-the-arts on three challenging benchmark datasets. 
We report mAP for all the compared models on the ILSVRC-2017 detection dataset. 
As can be seen from Table~\ref{table:imagenet}, the proposed approach performs better than the other compared models on most of the categories, achieving 1.9\% improvement against ZSDTD~\cite{Li_2019zeroshot}. 
The performance on the categories (\textit{i.e.} hamster, tiger, scorpion) which have similar concepts are much better than those (\textit{i.e.} ray, harmonica, maraca) which have less similar concepts. 
Moreover, our model has significant improvement in the syringe category from 3.1\% to 30.4\%, but the performance on the ray class is 25.5\% less than the first place. 
We note that the syringe has twenty-three similar categories in training dataset while the ray only has five. 
This phenomenon indicates that our approach can effectively transfer the knowledge to the unseen class domain when we have enough similar categories in seen class domain. 
While there are few similar classes in the training dataset, it is difficult for our generative model to synthesize discriminative visual features. 

For the MSCOCO and VG datasets, We report the results in terms of Recall@100 for all the compared methods in Table~\ref{table:coco_vg}, with three different IoU overlap thresholds (\textit{i.e}. 0.4, 0.5, 0.6). 
In addition, for convenient discussion, IoU 0.5 is used as an example in this paper unless specified otherwise. 
From the experimental results, we observe that the proposed model generally performs much better than the other compared baselines. 
Specifically, the proposed model increases Recall@100 to 44.6\% from 34.3\% achieved by the method~\cite{Li_2019zeroshot} on MSCOCO, and increases Recall@100 to 11.3\% from 7.2\% on VG.

\begin{figure*}[!tb]
\begin{center}
\includegraphics[width=\linewidth]{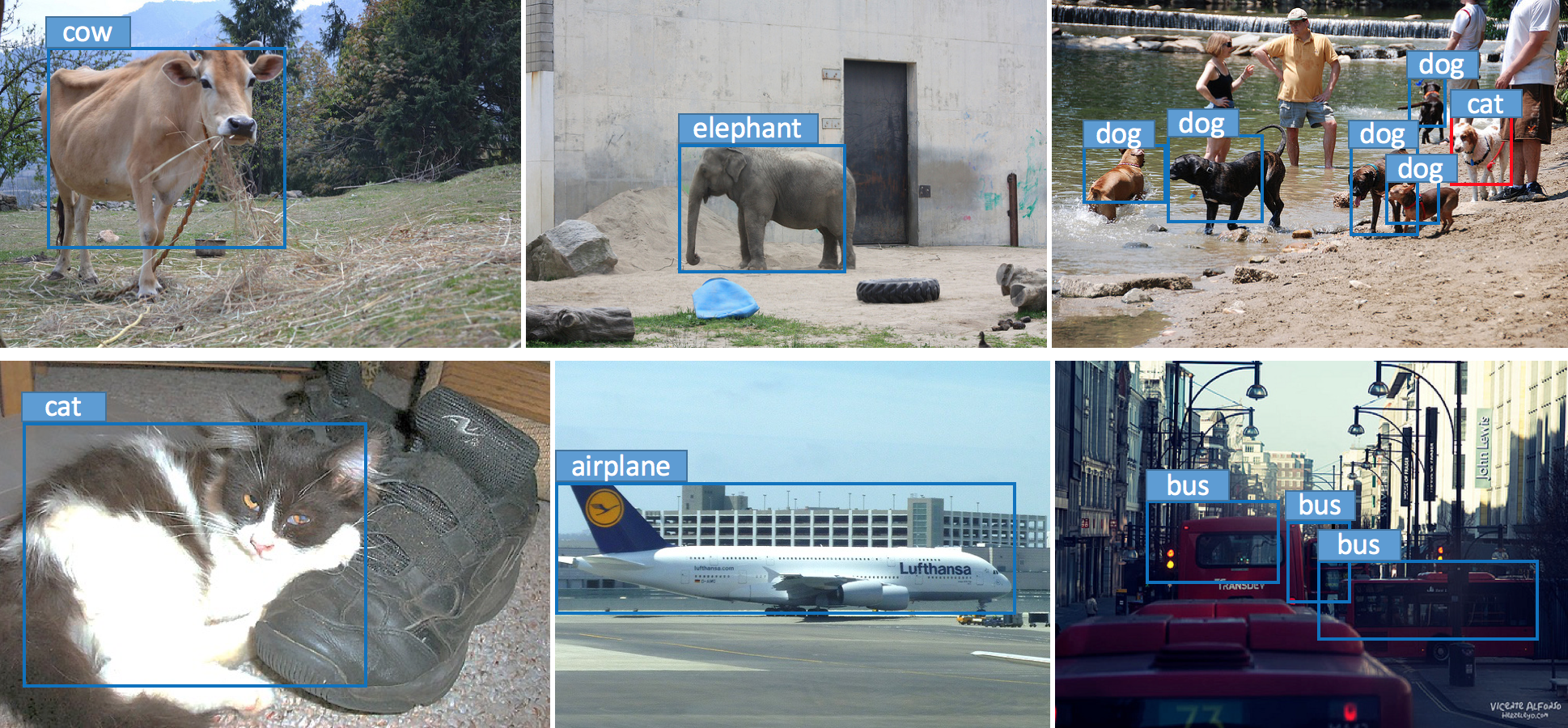}
   \caption{Some examples of zero-shot object detection using the proposed approach. 
   }
 \label{fig:quares}
\end{center}
\end{figure*}

\subsection{4.3. Ablation Studies}
We study the effects of different components in our model and report the experimental results. 
We still use mAP as the evaluation metric for ILSVRC and Recall@100 for MSCOCO and VG. 
We design a baseline which uses the traditional visual-semantic embedding approach. 
Note that in the inference stage, the network architectures of the baseline and our approach are the same except that our approach directly maps the RoI features into label space while the baseline maps the features into semantic space followed by a nearest neighbor search. 

\noindent {\bf Effectiveness of IoUGAN:} To study the effect of the IoU information and whether the synthesized background features are useful or not, We first report the results which use the features generated by only CFU with the background features sampling from the detection training dataset. 
Note that these generated features do not contains the IoU variance. 
Then we use FFU to add the IoU variance to the features from CFU. 
In the end, we use the full architecture of IoUGAN with the ability to synthesize the class-specific background features.
From the experimental results, as shown in Table~\ref{compon}, we can see that only using CFU, our approach can outperform the baseline by 8.9\%,  0.6\% and 3.8\% on three datasets, respectively. 
Moreover, adding the IoU variance by FFU will dramatically improve the performance of ZSD from 27.0\%, 3.6\% and 20.1\% to 39.2\%, 7.5\% and 24.5\%, which indicates the great importance of the IoU variance. 
The ZSD performance is further improved when using the synthesized background features which make the detector more discriminatory. 

\noindent {\bf Effectiveness of the Loss Functions:}
To study the effect of the loss functions, we first report results only using the Wasserstein loss. Then we add the classification loss $\mathcal{L}_{cls}$ and the embedding loss $\mathcal{L}_{emb}$, respectively. 
To the end, we report the results using all the loss functions. 
As shown in Table~\ref{loss_exp}, both the classification loss and the embedding loss can further improve the performance.
%

\subsection{4.4. Generalized Zero-Shot Object Detection}
We also show the results of the Generalized Zero-Shot Object Detection (GZSD) in Table~\ref{gzsl}. 
We compare to the baseline method presented in the ablation study and the Polarity Loss~\cite{rahman2018polarity}.  
The results in Table~\ref{gzsl} shows proposed GTNet performs better than the baseline and the Polarity Loss. 

\subsection{4.5. Qualitative Results}
Figure~\ref{fig:quares} shows the output detection by the proposed approach with the MSCOCO dataset being used as an example here.
These examples confirm that the proposed method can detect classes that are unseen during the training process. 
The failure detection tends to happen when objects are small and have similar classes in the set of unseen classes. 

\section{5. Conclusions}

In this paper, we propose GTNet, which consists of an Object Detection Module and a Knowledge Transfer Module, to tackle the ZSD problem.
%
%
GTNet naturally embeds a feature synthesizer in the Knowledge Transfer Module. 
In order to generate unseen class features with the IoU variance, we design a novel generative model IoUGAN as the feature synthesizer. 
IoUGAN is composed of three unit models (\textit{i.e.} CFU, FFU, and BFU).
The CFU aims to generate features with the intra-class variance for each unseen class.
FFU and BFU add the IoU variance to the results of CFU and output the class-specific foreground and background features, respectively.
We conducted extensive experiments in order to demonstrate the superiority of our model and investigated the effectiveness of different components.
In the future, we will investigate how to generate more discriminative features for the unseen classes that have few similar categories in the training dataset. 

\section*{Acknowledgment}
This work was supported by National Natural Science Foundation of China (No.61901184, No.61771329 and No. 61876210).

\clearpage
{
\bibliographystyle{aaai}
\bibliography{aaai}
}

\end{document}